\documentclass{article}

\usepackage{microtype}
\usepackage{booktabs} 
\usepackage{longtable}
\usepackage{graphicx}
\usepackage{subcaption} 
\usepackage{multirow}
\usepackage{multicol}
\usepackage{listings}
\usepackage{hyperref}

\usepackage[accepted]{icml2025}

\usepackage{amsmath}
\usepackage{amssymb}
\usepackage{mathtools}
\usepackage{amsthm}

\usepackage[capitalize,noabbrev]{cleveref}

\theoremstyle{plain}

\theoremstyle{definition}

\theoremstyle{remark}

\usepackage[textsize=tiny]{todonotes}

\icmltitlerunning{Foundation Models for Clinical Records at Health System Scale}

\begin{document}

\twocolumn[
\icmltitle{Foundation Models for Clinical Records at Health System Scale}



\icmlsetsymbol{equal}{*}

\begin{icmlauthorlist}
\icmlauthor{Haresh Rengaraj Rajamohan}{equal,nyu}
\icmlauthor{Xiang Gao}{equal,nyu}
\icmlauthor{Weicheng Zhu}{nyu}
\icmlauthor{Shih-Lun Huang}{nyu}
\icmlauthor{Long Chen}{nyu}
\icmlauthor{Kyunghyun Cho}{nyu}
\icmlauthor{Cem M. Deniz}{nyu}
\icmlauthor{Narges Razavian}{nyu}
\end{icmlauthorlist}

\icmlaffiliation{nyu}{New York University}

\icmlcorrespondingauthor{Haresh Rajamohan / Xiang Gao}{\{hrr288, x.gao\}@nyu.edu}
\icmlcorrespondingauthor{Narges Razavian}{narges.razavian@nyulangone.org}

\icmlkeywords{Machine Learning, ICML}

\vskip 0.3in
]



\printAffiliationsAndNotice{\icmlEqualContribution} 

\begin{abstract}
Large-scale pretraining has transformed modeling of language and other data types, but its potential remains underexplored in healthcare with structured electronic health records (EHRs). We present a novel generative pretraining strategy for sequential EHR data using next-visit event prediction. Our model learns to autoregressively generate various tokenized clinical events for the next visit based on patient history and inherently handles the joint prediction of heterogeneous data types. Additionally, we introduce regularization on predicting repeated events and highlight a key pitfall in EHR-based foundation model evaluations: repeated event tokens can inflate performance metrics when new onsets are not distinguished from subsequent occurrences. Our model is evaluated via zero-shot prediction for forecasting dementia and knee osteoarthritis incidence within 2 and 5 years, and the model performance rivals a fully fine-tuned masked pretrained Transformer baseline, demonstrating that our approach captures complex clinical dependencies without requiring costly task-specific fine-tuning.
\end{abstract}

\section{Introduction}

Early detection and progression forecasting for chronic conditions like dementia, osteoarthritis, and cancers can significantly improve healthcare outcomes and can further optimize clinical trial design \cite{dubois2015timely, zhu2024predicting, arnold2022current,karsdal2016disease}. Electronic Health Records (EHRs) offer an enormous amount of longitudinal data for this purpose but present challenges due to their sequential, high-dimensional, irregular, and heterogeneous nature \cite{nordo2019use,choi2017using,xiao2018opportunities,shickel2017deep}. Traditional approaches of building models for prediction of specific diseases within specific timeframes are resource-intensive, lack flexibility, and fail to leverage interrelations among various clinical outcomes. This motivates the development of generative foundation models that learn comprehensive representations from EHR data and generate patient trajectories.

Prior work in training foundation models on EHR data with masked pretraining or encoder-decoder architecture requires an additional fine-tuning stage for adapting the model to particular downstream tasks \cite{li2020behrt, yang2023transformehr}. This requires curating additional task-specific datasets. More recently, generative pretraining for EHRs has gained traction due to success in other data modalities and the flexibility for zero-shot inference \cite{mcdermott2023event, renc2024zero}. However, many of these approaches are based on the next-token paradigm and trained on specific clinical records like these from intensive care units (ICU), which fail to demonstrate the challenges we outline below in learning longitudinal records from hospital visits. A more comprehensive discussion of related work can be found in Appendix \ref{app:related_work}.

A key challenge in modeling longitudinal EHR data is that clinical events that occurred within a single patient visit lack fine-grained temporal order, making the standard next-token prediction insufficient. Similarly to \citet{steinberg2021language}, we employ a next-visit multi-label prediction paradigm, but we use a decoder-only Transformer to perform zero-shot inference instead of just representation learning. Moreover, unlike text tokens, recurring event tokens, especially chronic conditions, can reappear throughout a patient’s record. This can lead the model to fail to learn new onsets of diseases and can mask the model's true ability to detect emerging events during evaluations. \citet{kraljevic2022foresight} first showed that model performance degrades on predicting new concepts. We provide additional empirical evidence and improve predictions on new onsets through regularization.

\begin{figure*}[htbp]
\centering 

\includegraphics[width=0.85\textwidth]{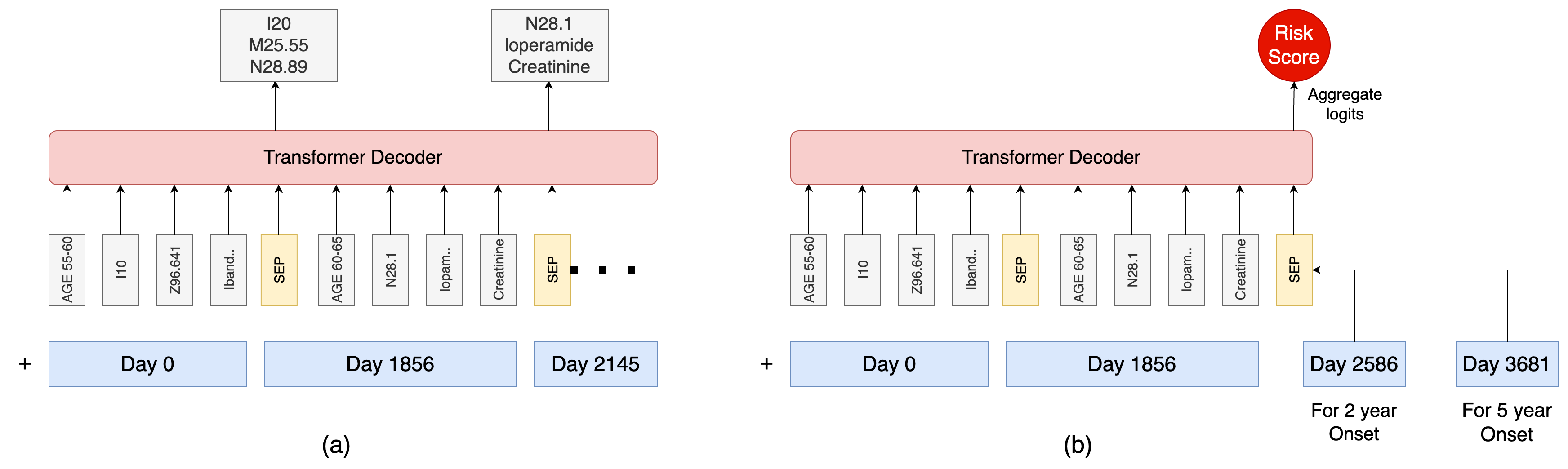} 
  \caption{Illustrative examples to outline the GPT-EHR system, showing both the pretraining setup (a) and the evaluation methodology (b).}
  \label{fig:main_gpt_ehr}
\end{figure*}

In this work, we introduce a generative pretraining framework using next-visit event prediction for longitudinal EHR data. Our model autoregressively generates the joint state of medications, labs, and diagnoses for the next encounter, conditioned on patient history and the time of the next visit. Also, we show how repeated event tokens can inflate model performance through pretraining evaluations and propose regularization for repeated events to encourage the learning of new events. Compared to a fully fine-tuned BERT-based foundation model baseline, we demonstrate strong zero-shot generalization on forecasting dementia and knee osteoarthritis with different prediction horizons.

\section{Methods}

We develop and validate our approaches using a de-identified longitudinal EHR dataset from NYU Langone Health, a large hospital system with a diverse patient population in New York City, encompassing both inpatient and outpatient visits over a ten-year period from January 2013 to January 2023. The selected cohort comprised records from approximately 1.29 million unique patients (N=1,288,242), characterized by a median of 21 visits per patient (mean: 37.76, range: 2-2123). The entire selected cohort was randomly split into training (70\%), validation (15\%), and test (15\%) sets on a patient level.

\subsection{Tokenization and Input Representation}

Patient records are tokenized into sequences of clinical events associated with specific visits, including demographics, age at visit, medications, diagnoses, and lab results. Continuous variables, such as age and lab values, were discretized into quantile-based bins. Each category and bin from different events forms a unique token, resulting in a vocabulary size $\lvert \mathcal{V} \rvert$ of 42337 unique tokens. On average, each visit contains 11.16 tokens, and each patient trajectory comprises 474.21 tokens. We group sequences of tokens for each patient by their visits chronologically and apply a special separator token, \texttt{<sep>}, to the end of each visit to explicitly define visit boundaries.

We use rotary positional embeddings (RoPE) from \citet{su2024roformer} and make tokens within the same visit share a positional embedding corresponding to the time elapsed since the initial visit. Crucially, the \texttt{<sep>} token concluding visit $i$ received the positional embedding of the next visit $i+1$, explicitly encoding the inter-visit time interval.

\subsection{Model Architecture}

We utilize a decoder-only Transformer based on the GPT-2 architecture and provide an overview of the pipeline including data representations in Figure \ref{fig:main_gpt_ehr} \cite{radford2018improving}. We modify the causal attention mechanism to allow all tokens \textit{within the same visit} to attend to each other, maintaining causality in time across visits. Thus, a token in visit $v$ attends to all tokens in any visit $v' \le v$. For training efficiency, we use sequence packing to have multiple patients in one training sequence if their tokens can fit the context length. Our implementation allows attention across concatenated patient sequences, potentially offering broader context alongside improved GPU throughput.

\subsection{Objective and Loss Formulation} 

The model is pretrained for next-visit multi-label prediction. Given patient history $H_{t_i}$ up to visit $i$ occurring at time $t_i$ and the time $t_{i+1}$ of the next visit (via the \texttt{<sep>} token's position), the model learns to predict the set of all tokens $\{k\}$ present in visit $i+1$. The prediction is generated based on the output representation of the \texttt{<sep>} token from visit $i$ which passed through a linear layer followed by a sigmoid activation to produce probabilities $\hat{P}_{i+1, k} = P(k | H_{t_i}, t_{i+1})$ for each token $k \in \mathcal{V}$. 

Furthermore, to encourage predicting the \textit{onset} of new clinical events, especially for chronic conditions with frequent token repetitions, we introduced a regularization scheme that, for each token, penalizes the loss according to its frequency in the patient's history $H_{t_i}$. Tokens appearing more often in the past received a lower weight in the loss calculation. Specifically, the weight is obtained through a power decay function that depends on the number of repeats and a hyperparameter $\delta$ that controls the strength of the regularization. The final loss is a weighted binary cross-entropy, and please refer to Appendix \ref{app:A_regularization_loss} for Equation \ref{eq:loss_mod_appendix} and the detailed formulation of the weighting mechanism.

\section{Experiments}

We perform large-scale pretraining of the proposed framework on our training sub-cohort. The final model size is close to $1.6$ billion parameters with a maximum context length of $512$ tokens; other hyperparameters can be found in Appendix \ref{app:_hyperparam}. The pretraining takes around $7$ days with two Nvidia A100 GPUs.  Similar to prior work, we start by examining the overall next-visit prediction performance for model development and ablation studies. Then, we demonstrate the clinical utility of our approach via zero-shot inference evaluations on separately curated data of predefined disease prediction tasks. We compare our approach to a foundation model with masked pretraining that has been fine-tuned for specific disease prediction tasks at specific timeframes. We show that our approach achieves similar or better performance without any additional task-specific fine-tuning.

\subsection{Pretraining Evaluation} 
To assess the pretraining process and the effect of the repeat token regularization, we evaluate the overall next-visit prediction performance on two specific conditions namely dementia and pancreatic cancer. For these evaluations, decision thresholds for identifying a predicted event are optimized using the F1 score on the validation set. These thresholds are then applied to the test set to calculate precision and recall. 

\begin{figure}[H]
\begin{center}
\centerline{\includegraphics[width=\columnwidth]{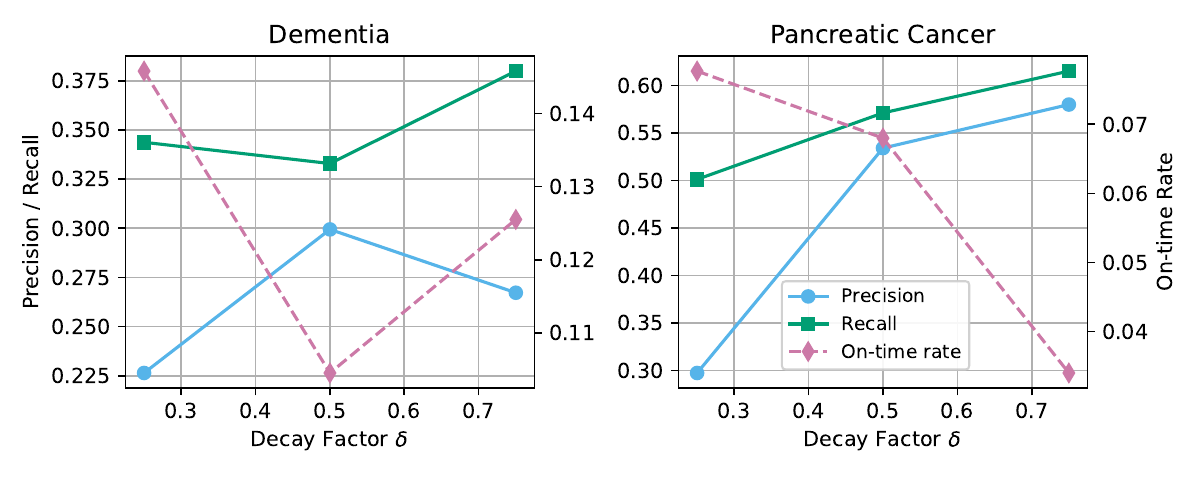}}
\vskip -0.15in
\caption{Effect of decay regularization on repeated tokens. Showing precision/recall against the primary (left) y-axis, while dashed lines show the on-time rate against the secondary (right) y-axis, for distinct decay factors. Left: dementia. Right: pancreatic cancer.}
\label{fig:on-time}
\end{center}
\vskip -0.4in
\end{figure}

Importantly, we define a new metric, the \textit{on-time rate} $r=\text{TP}_{\leq t_{1}}/\text{TP}_{total}$, which measures the proportion of true positive predictions that occur at or before the first recorded onset of the condition in the patient trajectory. Here, true positives are defined at the trajectory level: a patient is considered a true positive if they eventually develop the condition and the model predicts it at any point across the rolling evaluation windows. That is, if any prediction window for the patient correctly flags the future onset, the patient counts as a true positive. The on-time rate helps us to distinguish performance on forecasting new onsets versus merely repeating known information. 
For calculating precision and recall, we only count the prediction as correct if it appears in the next visit, i.e., we do not relax the evaluation by having a time range over future visits that can contain the correct predictions, as in some prior work \cite{kraljevic2022foresight}.

We investigated the effects of our decay regularization approach on repeated clinical events and varied the decay factor parameter $\delta$ from $0$ to $1$ where smaller $\delta$ leads to larger penalization. As illustrated in Figure \ref{fig:on-time}, increasing the penalization (smaller $\delta$) significantly improved the on-time rate at the expense of overall precision and recall. For dementia, a decay factor of $0.25$ yielded the highest on-time rate, despite the reduced overall precision and recall, both of which declined with increased penalization (not strictly). For pancreatic cancer, this trend becomes more apparent as the highest on-time rate was achieved with the strongest regularization, which manifests as lower overall precision and recall. We selected $\delta = 0.5$ for all downstream experiments as a balanced operating point. This highlights an inherent trade-off: prioritizing new event prediction can impact overall repetition-based metrics.

\subsection{Zero-shot Inference}\label{sec:zero_shot_eval}

Without task-specific fine-tuning, we evaluated the zero-shot ability of our pretrained model to forecast future disease incidence for dementia and knee osteoarthritis (OA). For each condition, we set specific prediction tasks along with the corresponding curated data: forecasting the first incidence within the (i) next 2 years and (ii) the next 5 years (two tasks). The ground truth label was defined by the first occurrence of the condition, defined as a set of diagnosis and medication codes within the prediction window. Details of the task setup can be found in Appendix \ref{app:A_zeroshot_details}.

\begin{figure}[H]
\begin{center}
\centerline{\includegraphics[width=\columnwidth]{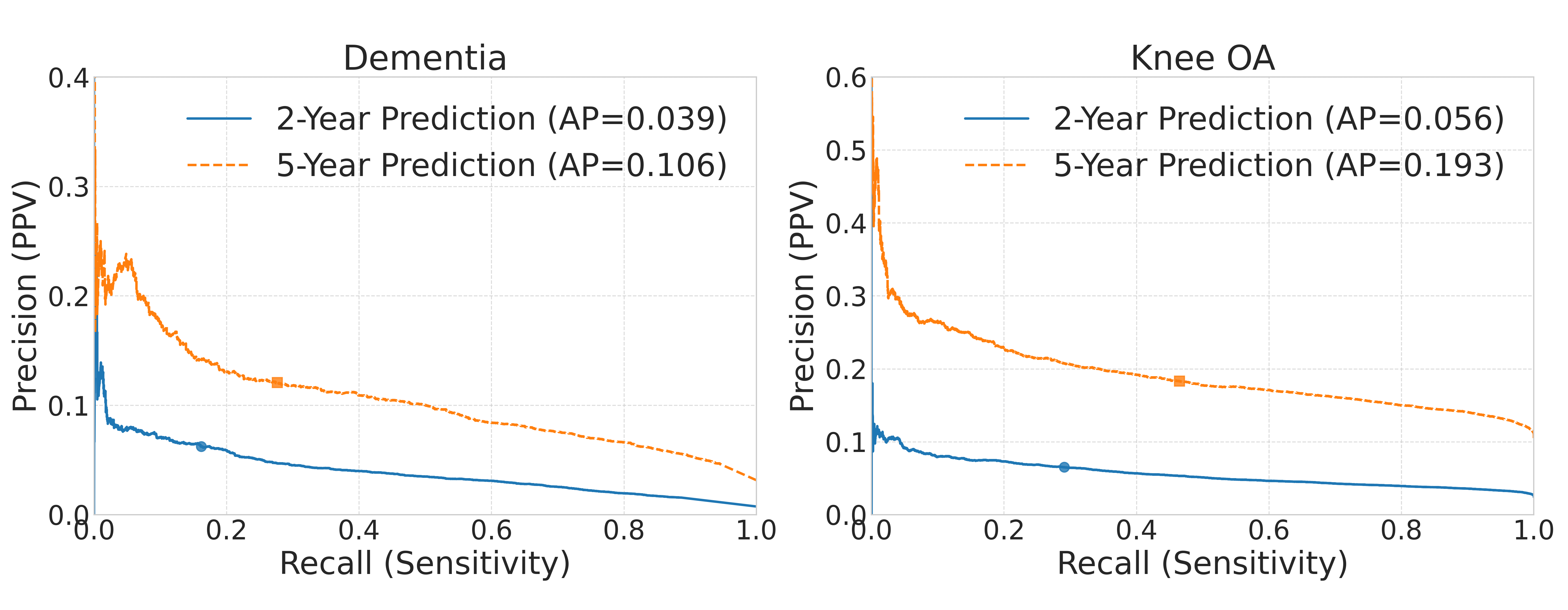}}
\vskip -0.15in
\caption{Performance of our model on long-term risk prediction, illustrated with Precision-Recall curves for: (a) Dementia, comparing 2-year (solid blue line) and 5-year (dashed orange line) prediction horizons; and (b) Knee Osteoarthritis (OA), with the same prediction horizons. Average precision are displayed for each curve in the legend. Optimal threshold points are indicated by markers (circle for 2-year, square for 5-year) on each curve.}
\label{fig:pr_curves}
\end{center}
\vskip -0.32in
\end{figure}

To ensure the model is predicting future new onset based on prior history, rather than simply repeating the chronic condition if already present, we applied strict filtering criteria on the input and output window when curating this data. We exclude patients' windows if their input history windows contain any label code or the disease onset occurred within one year following the prediction time point. This ensures the model forecasts medium- to long-term risk rather than imminent events and reduces the risks of label leakage.

For dementia prediction, our generative pretrained model achieved an AUROC of 0.814 at the 2-year horizon, outperforming the fine-tuned BERT baseline as shown in Table \ref{tab:combined_onset_prediction}. Similarly, at the 5-year horizon, our GPT-based model maintained superior predictive performance compared to the baseline. Also, our model obtained slightly lower precision compared to BERT, and we present the precision-recall curve in Figure \ref{fig:pr_curves}. We chose the conditions not only because they are of great clinical importance but also  because they exhibit real world challenges that have been ignored in most prior work. As an additional baseline, we also compared the zero-shot prediction rates of LLaMA-3.3-70B large language model, under the exact same task setup. The LLM did not obtain comparable performance, and further information can be found in Appendix \ref{app:llm} .

\begin{table}[htbp]
\centering

\resizebox{0.48\textwidth}{!}{%
\begin{tabular}{@{}lllcc@{}} 
  \toprule
  Condition & Time Horizon & Approach         & AUROC                  & AUPRC                  \\
  \midrule
  \multirow{4}{*}{Dementia} & \multirow{2}{*}{2 years} & GPT Zero-shot    & 0.814 (0.804, 0.823)   & 0.039 (0.036, 0.044)   \\
                            &                          & Fine-tuned BERT  & 0.731 (0.730, 0.733)    & 0.050 (0.050, 0.051)     \\
                            \cmidrule(lr){2-5} 
                            & \multirow{2}{*}{5 years} & GPT Zero-shot    & 0.781 (0.770, 0.790)     & 0.106 (0.096, 0.116)   \\
                            &                          & Fine-tuned BERT  & 0.721 (0.720, 0.722)    & 0.146 (0.145, 0.147)   \\
  \midrule
  \multirow{4}{*}{Knee OA}  & \multirow{2}{*}{2 years} & GPT Zero-shot    & 0.724 (0.719, 0.730)    & 0.056 (0.053, 0.059)   \\
                            &                          & Fine-tuned BERT  & 0.743 (0.743, 0.744)   & 0.064 (0.064, 0.064)   \\
                            \cmidrule(lr){2-5} 
                            & \multirow{2}{*}{5 years} & GPT Zero-shot    & 0.693 (0.686, 0.699)   & 0.193 (0.185, 0.203)   \\
                            &                          & Fine-tuned BERT  & 0.717 (0.716, 0.718)   & 0.220 (0.219, 0.220)     \\
  \bottomrule
\end{tabular}%
} 

\caption{Comparison of prediction metrics on dementia and knee OA onset using zero-shot and fine-tuned models across different time horizons with 95\% confidence intervals.}
\label{tab:combined_onset_prediction}
\end{table}

For knee OA prediction, the fine-tuned BERT baseline exhibited slightly superior AUROC performance (2-year: 0.743, 5-year: 0.717) compared to our GPT-based model (2-year: 0.724, 5-year: 0.693). Nonetheless, the GPT-based model still performed strongly given its zero-shot nature, showing its utility in scenarios lacking task-specific annotations and fine-tuning resources. Overall, these results demonstrate that our generative pretraining approach can robustly generalize to diverse predictive tasks across varying disease progression timelines.

\section{Conclusion and Discussion}

In this work, we present a new generative pretraining framework for sequential EHR data, centered on next-visit multi-label event prediction with regularization on repeated clinical events. Our proposed model demonstrates excellent zero-shot inference capabilities. Notably, its performance in predicting the 2-year and 5-year incidence of dementia and knee OA matches, and in some cases exceeds, that of fully fine-tuned masked pretrained Transformer baselines. This highlights that our approach can learn and zero-shot generalize directly from sequential EHR data, without the need for costly task-specific annotations and training.

One key contribution of this work is to highlight an important lesson in evaluations for foundation models trained on longitudinal EHRs. Our pretraining evaluations revealed an inherent trade-off when emphasizing the prediction of new clinical events: stronger regularization improved the on-time rate for new onsets but could reduce overall precision and recall. This underscores that if models are blindly evaluated and optimized solely on certain metrics without careful design, there is a risk of developing clinically ineffective systems that primarily repeat patient history rather than effectively predicting new disease onsets. 

For instance, a model might achieve high recall by simply echoing chronic diagnoses seen earlier in the trajectory, without demonstrating a genuine understanding of when a condition first arose or changed in severity. Our regularization scheme directly mitigates this issue by penalizing predictions of frequently repeated clinical events, thereby guiding the model to prioritize meaningful predictions of new disease occurrences. By explicitly separating and evaluating these two phenomena - predicting new onsets versus repeating known information - evaluations can better reflect real-world clinical needs, such as providing early warnings for new diagnoses or enabling timely intervention. This approach ensures more clinically relevant and accurate assessments of model performance.

While outpatient EHR data inherently lacks comprehensive details on patient lifestyle, socioeconomic status, or environmental exposures, our work demonstrates the predictive value is still extractable from the recorded clinical trajectory. Many conditions, particularly those with slower progression like knee osteoarthritis or dementia, leave discernible traces in the form of evolving symptoms, lab abnormalities, or precursor diagnoses within the EHR over time. Our findings, particularly the zero-shot results, indicate that the autoregressive learning framework is adept at identifying these temporal patterns and early warning signs. 

\textbf{Limitations and Future Work.} We have not fully leveraged the generative nature of this pretrained model as we have only focused on one-step predictions so far, and it is not entirely clear what is the best way to sequentially generate future trajectories with a next-visit prediction framework. In contrast, for large language models, there is a rich literature and methods on inference-time algorithms and scaling such as improving reasoning and generating long-form responses through reinforcement learning, and we believe these methods could potentially continue to improve foundation models on sequential tabular data. Also, even though we have access to a large-scale EHR dataset, the scale of the training data still falls short of foundation models on other modalities like language and vision. Future research remains on designing data-efficient algorithms and synthetic data generation.

\bibliography{ref}

\begin{thebibliography}{26}
\providecommand{\natexlab}[1]{#1}
\providecommand{\url}[1]{\texttt{#1}}
\expandafter\ifx\csname urlstyle\endcsname\relax
  \providecommand{\doi}[1]{doi: #1}\else
  \providecommand{\doi}{doi: \begingroup \urlstyle{rm}\Url}\fi

\bibitem[Arnold et~al.(2022)Arnold, Morgan, Rumgay, Mafra, Singh, Laversanne, Vignat, Gralow, Cardoso, Siesling, et~al.]{arnold2022current}
Arnold, M., Morgan, E., Rumgay, H., Mafra, A., Singh, D., Laversanne, M., Vignat, J., Gralow, J.~R., Cardoso, F., Siesling, S., et~al.
\newblock Current and future burden of breast cancer: Global statistics for 2020 and 2040.
\newblock \emph{The Breast}, 66:\penalty0 15--23, 2022.

\bibitem[Choi et~al.(2017)Choi, Schuetz, Stewart, and Sun]{choi2017using}
Choi, E., Schuetz, A., Stewart, W.~F., and Sun, J.
\newblock Using recurrent neural network models for early detection of heart failure onset.
\newblock \emph{Journal of the American Medical Informatics Association}, 24\penalty0 (2):\penalty0 361--370, 2017.

\bibitem[Devlin et~al.(2019)Devlin, Chang, Lee, and Toutanova]{devlin2019bert}
Devlin, J., Chang, M.-W., Lee, K., and Toutanova, K.
\newblock Bert: Pre-training of deep bidirectional transformers for language understanding.
\newblock In \emph{Proceedings of the 2019 conference of the North American chapter of the association for computational linguistics: human language technologies, volume 1 (long and short papers)}, pp.\  4171--4186, 2019.

\bibitem[Dubois et~al.(2015)Dubois, Padovani, Scheltens, Rossi, and Dell’Agnello]{dubois2015timely}
Dubois, B., Padovani, A., Scheltens, P., Rossi, A., and Dell’Agnello, G.
\newblock Timely diagnosis for alzheimer’s disease: a literature review on benefits and challenges.
\newblock \emph{Journal of Alzheimer’s disease}, 49\penalty0 (3):\penalty0 617--631, 2015.

\bibitem[Fallahpour et~al.(2024)Fallahpour, Alinoori, Ye, Cao, Afkanpour, and Krishnan]{fallahpour2024ehrmamba}
Fallahpour, A., Alinoori, M., Ye, W., Cao, X., Afkanpour, A., and Krishnan, A.
\newblock Ehrmamba: Towards generalizable and scalable foundation models for electronic health records.
\newblock \emph{arXiv preprint arXiv:2405.14567}, 2024.

\bibitem[Guo et~al.(2023)Guo, Steinberg, Fleming, Posada, Lemmon, Pfohl, Shah, Fries, and Sung]{guo2023ehr}
Guo, L.~L., Steinberg, E., Fleming, S.~L., Posada, J., Lemmon, J., Pfohl, S.~R., Shah, N., Fries, J., and Sung, L.
\newblock Ehr foundation models improve robustness in the presence of temporal distribution shift.
\newblock \emph{Scientific Reports}, 13\penalty0 (1):\penalty0 3767, 2023.

\bibitem[Guo et~al.(2024)Guo, Fries, Steinberg, Fleming, Morse, Aftandilian, Posada, Shah, and Sung]{guo2024multi}
Guo, L.~L., Fries, J., Steinberg, E., Fleming, S.~L., Morse, K., Aftandilian, C., Posada, J., Shah, N., and Sung, L.
\newblock A multi-center study on the adaptability of a shared foundation model for electronic health records.
\newblock \emph{NPJ Digital Medicine}, 7\penalty0 (1):\penalty0 171, 2024.

\bibitem[Johnson et~al.(2020)Johnson, Bulgarelli, Pollard, Horng, Celi, and Mark]{johnson2020mimic}
Johnson, A., Bulgarelli, L., Pollard, T., Horng, S., Celi, L.~A., and Mark, R.
\newblock Mimic-iv.
\newblock \emph{PhysioNet. Available online at: https://physionet. org/content/mimiciv/1.0/(accessed August 23, 2021)}, pp.\  49--55, 2020.

\bibitem[Karsdal et~al.(2016)Karsdal, Michaelis, Ladel, Siebuhr, Bihlet, Andersen, Guehring, Christiansen, Bay-Jensen, and Kraus]{karsdal2016disease}
Karsdal, M., Michaelis, M., Ladel, C., Siebuhr, A., Bihlet, A., Andersen, J., Guehring, H., Christiansen, C., Bay-Jensen, A., and Kraus, V.
\newblock Disease-modifying treatments for osteoarthritis (dmoads) of the knee and hip: lessons learned from failures and opportunities for the future.
\newblock \emph{Osteoarthritis and cartilage}, 24\penalty0 (12):\penalty0 2013--2021, 2016.

\bibitem[Kraljevic et~al.(2022)Kraljevic, Bean, Shek, Bendayan, Hemingway, Yeung, Deng, Baston, Ross, Idowu, et~al.]{kraljevic2022foresight}
Kraljevic, Z., Bean, D., Shek, A., Bendayan, R., Hemingway, H., Yeung, J.~A., Deng, A., Baston, A., Ross, J., Idowu, E., et~al.
\newblock Foresight--generative pretrained transformer (gpt) for modelling of patient timelines using ehrs.
\newblock \emph{arXiv preprint arXiv:2212.08072}, 2022.

\bibitem[Li et~al.(2020)Li, Rao, Solares, Hassaine, Ramakrishnan, Canoy, Zhu, Rahimi, and Salimi-Khorshidi]{li2020behrt}
Li, Y., Rao, S., Solares, J. R.~A., Hassaine, A., Ramakrishnan, R., Canoy, D., Zhu, Y., Rahimi, K., and Salimi-Khorshidi, G.
\newblock Behrt: transformer for electronic health records.
\newblock \emph{Scientific reports}, 10\penalty0 (1):\penalty0 7155, 2020.

\bibitem[Li et~al.(2022)Li, Mamouei, Salimi-Khorshidi, Rao, Hassaine, Canoy, Lukasiewicz, and Rahimi]{li2022hi}
Li, Y., Mamouei, M., Salimi-Khorshidi, G., Rao, S., Hassaine, A., Canoy, D., Lukasiewicz, T., and Rahimi, K.
\newblock Hi-behrt: hierarchical transformer-based model for accurate prediction of clinical events using multimodal longitudinal electronic health records.
\newblock \emph{IEEE journal of biomedical and health informatics}, 27\penalty0 (2):\penalty0 1106--1117, 2022.

\bibitem[McDermott et~al.(2023)McDermott, Nestor, Argaw, and Kohane]{mcdermott2023event}
McDermott, M., Nestor, B., Argaw, P., and Kohane, I.~S.
\newblock Event stream gpt: a data pre-processing and modeling library for generative, pre-trained transformers over continuous-time sequences of complex events.
\newblock \emph{Advances in Neural Information Processing Systems}, 36:\penalty0 24322--24334, 2023.

\bibitem[Nordo et~al.(2019)Nordo, Levaux, Becnel, Galvez, Rao, Stem, Prakash, and Kush]{nordo2019use}
Nordo, A.~H., Levaux, H.~P., Becnel, L.~B., Galvez, J., Rao, P., Stem, K., Prakash, E., and Kush, R.~D.
\newblock Use of ehrs data for clinical research: historical progress and current applications.
\newblock \emph{Learning health systems}, 3\penalty0 (1):\penalty0 e10076, 2019.

\bibitem[Pang et~al.(2021)Pang, Jiang, Kalluri, Spotnitz, Chen, Perotte, and Natarajan]{pang2021cehr}
Pang, C., Jiang, X., Kalluri, K.~S., Spotnitz, M., Chen, R., Perotte, A., and Natarajan, K.
\newblock Cehr-bert: Incorporating temporal information from structured ehr data to improve prediction tasks.
\newblock In \emph{Machine Learning for Health}, pp.\  239--260. PMLR, 2021.

\bibitem[Radford et~al.(2018)Radford, Narasimhan, Salimans, Sutskever, et~al.]{radford2018improving}
Radford, A., Narasimhan, K., Salimans, T., Sutskever, I., et~al.
\newblock Improving language understanding by generative pre-training.
\newblock 2018.

\bibitem[Rasmy et~al.(2021)Rasmy, Xiang, Xie, Tao, and Zhi]{rasmy2021med}
Rasmy, L., Xiang, Y., Xie, Z., Tao, C., and Zhi, D.
\newblock Med-bert: pretrained contextualized embeddings on large-scale structured electronic health records for disease prediction.
\newblock \emph{NPJ digital medicine}, 4\penalty0 (1):\penalty0 86, 2021.

\bibitem[Renc et~al.(2024)Renc, Jia, Samir, Was, Li, Bates, and Sitek]{renc2024zero}
Renc, P., Jia, Y., Samir, A.~E., Was, J., Li, Q., Bates, D.~W., and Sitek, A.
\newblock Zero shot health trajectory prediction using transformer.
\newblock \emph{NPJ Digital Medicine}, 7\penalty0 (1):\penalty0 256, 2024.

\bibitem[Shang et~al.(2019)Shang, Ma, Xiao, and Sun]{shang2019pre}
Shang, J., Ma, T., Xiao, C., and Sun, J.
\newblock Pre-training of graph augmented transformers for medication recommendation.
\newblock \emph{arXiv preprint arXiv:1906.00346}, 2019.

\bibitem[Shickel et~al.(2017)Shickel, Tighe, Bihorac, and Rashidi]{shickel2017deep}
Shickel, B., Tighe, P.~J., Bihorac, A., and Rashidi, P.
\newblock Deep ehr: a survey of recent advances in deep learning techniques for electronic health record (ehr) analysis.
\newblock \emph{IEEE journal of biomedical and health informatics}, 22\penalty0 (5):\penalty0 1589--1604, 2017.

\bibitem[Steinberg et~al.(2021)Steinberg, Jung, Fries, Corbin, Pfohl, and Shah]{steinberg2021language}
Steinberg, E., Jung, K., Fries, J.~A., Corbin, C.~K., Pfohl, S.~R., and Shah, N.~H.
\newblock Language models are an effective representation learning technique for electronic health record data.
\newblock \emph{Journal of biomedical informatics}, 113:\penalty0 103637, 2021.

\bibitem[Su et~al.(2024)Su, Ahmed, Lu, Pan, Bo, and Liu]{su2024roformer}
Su, J., Ahmed, M., Lu, Y., Pan, S., Bo, W., and Liu, Y.
\newblock Roformer: Enhanced transformer with rotary position embedding.
\newblock \emph{Neurocomputing}, 568:\penalty0 127063, 2024.

\bibitem[Wornow et~al.(2023)Wornow, Thapa, Steinberg, Fries, and Shah]{wornow2023ehrshot}
Wornow, M., Thapa, R., Steinberg, E., Fries, J., and Shah, N.
\newblock Ehrshot: An ehr benchmark for few-shot evaluation of foundation models.
\newblock \emph{Advances in Neural Information Processing Systems}, 36:\penalty0 67125--67137, 2023.

\bibitem[Xiao et~al.(2018)Xiao, Choi, and Sun]{xiao2018opportunities}
Xiao, C., Choi, E., and Sun, J.
\newblock Opportunities and challenges in developing deep learning models using electronic health records data: a systematic review.
\newblock \emph{Journal of the American Medical Informatics Association}, 25\penalty0 (10):\penalty0 1419--1428, 2018.

\bibitem[Yang et~al.(2023)Yang, Mitra, Liu, Berlowitz, and Yu]{yang2023transformehr}
Yang, Z., Mitra, A., Liu, W., Berlowitz, D., and Yu, H.
\newblock Transformehr: transformer-based encoder-decoder generative model to enhance prediction of disease outcomes using electronic health records.
\newblock \emph{Nature communications}, 14\penalty0 (1):\penalty0 7857, 2023.

\bibitem[Zhu et~al.(2024)Zhu, Tang, Zhang, Rajamohan, Huang, Ma, Chaudhari, Madaan, Almahmoud, Chopra, et~al.]{zhu2024predicting}
Zhu, W., Tang, H., Zhang, H., Rajamohan, H.~R., Huang, S.-L., Ma, X., Chaudhari, A., Madaan, D., Almahmoud, E., Chopra, S., et~al.
\newblock Predicting risk of alzheimer’s diseases and related dementias with ai foundation model on electronic health records.
\newblock \emph{medRxiv}, 2024.

\end{thebibliography}
\bibliographystyle{icml2025}

\newpage
\appendix
\onecolumn
\textbf{\large Appendix}

\section{Related Work}
\label{app:related_work}

Previous works on foundation models in EHR have utilized masked prediction \cite{devlin2019bert} as the pretraining task in order to learn the relevant relationships. \citet{li2020behrt} employed masked pre-training with a Transformer architecture on sequences of standardized diagnosis codes (301 types) combined with age and positional data from patient EHRs. Fine-tuning for predicting future diagnoses significantly outperformed prior deep learning baselines. \citet{shang2019pre} proposed G-BERT, which integrates graph neural networks to capture medical code hierarchies and a BERT-based Transformer, pre-trained on single-visit EHR data, to improve medication recommendation. \citet{rasmy2021med} also pretrained on diagnosis codes from 28.5 million patients and demonstrated improved performance on downstream disease prediction tasks, particularly in the few-shot regime. Temporal information handling in BERT-based models evolved with strategies like artificial time tokens \cite{pang2021cehr} and hierarchical transformers for longer sequences with diverse medical concepts \cite{li2022hi}.

Inspired by the success of Generative Pretraining (GPT) and Large Language Models (LLMs), recent research has increasingly focused on adapting these pretraining techniques for EHRs \cite{radford2018improving}. \citet{yang2023transformehr} built TransformerEHR using a generative encoder-decoder architecture pretrained on EHR sequences. The pretraining objective was to predict all ICD codes of the next chronological visit in a sequence-to-sequence fashion, incorporating relative time differences between visits via input embeddings. \citet{kraljevic2022foresight} presented Foresight, a GPT-2 model trained on patient timelines represented as sequences of SNOMED concepts alongside dedicated tokens for clinical demographics and temporal markers, aiming to predict the next token in the sequence. \citet{renc2024zero} used clinical data from the MIMIC-IV \cite{johnson2020mimic}, with a GPT-style transformer trained with next-token prediction. They tokenized clinical events, quantized numerical data, and explicit time intervals as distinct tokens and performed zero-shot prediction of clinical outcomes. To handle long sequences efficiently, \citet{fallahpour2024ehrmamba} employed the state-space based Mamba architecture on the MIMIC IV dataset. They introduced a novel efficient multitask finetuning approach where prepended task-specific tokens serve as prompts, enabling a single model to address multiple downstream tasks. 

A distinct generative approach, predicting the set of medical codes for the next visit rather than a single next token, was proposed by \citet{steinberg2021language} (CLIMBR) using a GRU, showing learned representations served as effective features for downstream tasks via linear probing. This line of work was extended with next token prediction to show robustness to temporal shifts \cite{guo2023ehr} and adaptability to external data with limited continual pretraining \cite{guo2024multi}. \citet{wornow2023ehrshot} introduced EHRSHOT, a benchmark dataset containing longitudinal, de-identified structured EHR data from 6739 patients. EHRSHOT provides broader patient timelines as it is not restricted to only ICU or emergency department visits, unlike many prior EHR datasets.

Our work builds on the next-visit multi-label prediction paradigm. However, we utilize a decoder-only Transformer and introduce two key distinctions: (1) explicit conditioning on the next visit's timeline during training, enabling flexible zero-shot forecasting across various future horizons without retraining, and (2) a regularization to mitigate the impact of frequently repeated clinical events, promoting the prediction of new events. We evaluate this on a large-scale institutional clinical dataset covering both inpatient and outpatient data over 10 years, using rigorous benchmarks.

\section{Implementation Details}
\label{app:A_implementation}

\subsection{Tokenization and Input Representation Details}
\label{app:A_tokenization_details}
Patient records were processed into sequences of clinical events. Included event types were: patient demographics (self-reported ethnicity, race, sex), age at the time of visit, prescribed medications (mapped to RxNorm concepts), diagnoses (ICD-10 codes), and laboratory results (LOINC codes and values). Continuous variables, namely age and laboratory test results, were discretized into quantile-based bins (e.g., 10 bins for lab results). Each unique demographic category, discretized age bin, medication concept, ICD-10 code, and discretized lab result bin was treated as a distinct token. This resulted in a total vocabulary size $\lvert V \rvert$ of 42337 unique tokens. On average, each visit contained 11.16 tokens, and each patient trajectory comprised 474.21 tokens (median: 191).

\subsection{Regularization on Repeated Tokens: Loss Formulation}
\label{app:A_regularization_loss}
The model was pretrained on the task of next-visit prediction using a multi-label binary classification objective. Given a patient's history $H_{t_i}$ up to visit $i$ occurring at time $t_i$, and the time $t_{i+1}$ of the next visit (implicitly provided via the \texttt{<sep>} token's positional embedding), the model learns to predict the set of all tokens $\{k\}$ present in visit $i+1$. The prediction is generated based on the output representation of the \texttt{<sep>} token corresponding to visit $i$. This output is passed through a linear layer followed by a sigmoid activation to produce probabilities $\hat{P}_{i+1, k} = P(k | H_{t_i}, t_{i+1})$ for each token $k \in \mathcal{V}$. Let $V_{i+1}$ be the true multi-hot vector for visit $i+1$, where $V_{i+1, k}=1$ if token $k$ is present, and 0 otherwise.

To explicitly encourage the model to predict the \textit{onset} of new clinical events, we introduced a regularization scheme that penalizes the loss associated with repeated tokens. For each token $k$ that actually occurs in the target next visit $V_{i+1}$ (i.e., where $V_{i+1, k}=1$), we calculate a weight $w_{i+1, k}$ based on its count $c(k, H_{t_i})$ in the preceding history $H_{t_i}$. The weight $w_{i+1, k}$ is:
$$w_{i+1, k} = \max(\delta^{c(k, H_{t_i})}, w_{min})$$
where $\delta$ is a decay factor hyperparameter ($0 < \delta \le 1$) and $w_{min}$ is a minimum weight hyperparameter ($0 \le w_{min} < 1$). For token predictions where the ground truth is negative ($V_{i+1, k}=0$), the weight is 1. A decay factor $\delta < 1$ reduces the weight for tokens that have appeared more often. For example, if $\delta=0.5$, the weight for the first occurrence ($c=0$) is $\max(0.5^0, w_{min}) = 1$, for the second ($c=1$) it is $\max(0.5^1, w_{min}) = \max(0.5, w_{min})$, floored at $w_{min}$. The final loss is the weighted binary cross-entropy:

\begin{equation}
\begin{split}
    \mathcal{L}_{mod}(V_{i+1}, \hat{P}_{i+1}, H_{t_i}) = - \sum_{k=1}^{|\mathcal{V}|} & w'_{i+1, k} \Big[ V_{i+1, k} \log(\hat{P}_{i+1, k}) \\
    + (1 - V_{i+1, k}) \log(1 - \hat{P}_{i+1, k}) \Big]
\end{split}
\label{eq:loss_mod_appendix} 
\end{equation}
where $w'_{i+1, k} = w_{i+1, k}$ if $V_{i+1, k}=1$, and $w'_{i+1, k}=1$ if $V_{i+1, k}=0$. This nudges the model to focus on potentially novel events rather than highly predictable recurrent tokens.

\subsection{Evaluation Details}
\label{app:A_zeroshot_details}
For zero-shot evaluation, disease onset (label) was defined based on the first occurrence of any code from a predefined set of relevant diagnosis codes (ICD-10) and potentially medication codes associated with that condition within the specified 2- or 5-year prediction window. For instance, the dementia label relied on a group of specific ICD codes (e.g., G30.x, F01.x, F03.x) and dementia-related medication codes (e.g., RxNorm codes for donepezil, memantine). Similar specific code sets were defined for Knee Osteoarthritis. The model's output probabilities for tokens in the target condition's code set were aggregated (sum of logits) into a single probability.

\begin{longtable}{llll}
\caption{Diagnosis and medication codes for disease labeling.} \\
\toprule
\textbf{Disease} & \textbf{Type} & \textbf{Description} & \textbf{Code} \\
\midrule
\endfirsthead

\multicolumn{4}{l}{\textit{(Continued from previous page)}} \\
\toprule
\textbf{Disease} & \textbf{Type} & \textbf{Description} & \textbf{Code} \\
\midrule
\endhead

\midrule
\multicolumn{4}{r}{\textit{(Continued on next page)}} \\
\endfoot

\bottomrule
\endlastfoot

Dementia & Diagnosis & Vascular dementia without behavioral disturbance & F01.50 \\
Dementia & Diagnosis & Vascular dementia with behavioral disturbance & F01.51 \\
Dementia & Diagnosis & Dementia in other diseases w/o behavioral disturbance & F02.80 \\
Dementia & Diagnosis & Dementia in other diseases w/ behavioral disturbance & F02.81 \\
Dementia & Diagnosis & Unspecified dementia without behavioral disturbance & F03.90 \\
Dementia & Diagnosis & Unspecified dementia with behavioral disturbance & F03.91 \\
Dementia & Diagnosis & Amnestic disorder due to physiological condition & F04 \\
Dementia & Diagnosis & Progressive supranuclear ophthalmoplegia & G23.1 \\
Dementia & Diagnosis & Alzheimer's disease with early onset & G30.0 \\
Dementia & Diagnosis & Alzheimer's disease with late onset & G30.1 \\
Dementia & Diagnosis & Other Alzheimer's disease & G30.8 \\
Dementia & Diagnosis & Alzheimer's disease, unspecified & G30.9 \\
Dementia & Diagnosis & Pick's disease & G31.01 \\
Dementia & Diagnosis & Other frontotemporal dementia & G31.09 \\
Dementia & Diagnosis & Senile degeneration of brain & G31.1 \\
Dementia & Diagnosis & Dementia with Lewy bodies & G31.83 \\
Dementia & Diagnosis & Mild cognitive impairment & G31.84 \\
Dementia & Diagnosis & Corticobasal degeneration & G31.85 \\
Dementia & Diagnosis & Degenerative disease of nervous system, unspecified & G31.9 \\
Dementia & Medication & RIVASTIGMINE TARTRATE 1.5 MG ORAL CAP & 57619 \\
Dementia & Medication & MEMANTINE 10 MG ORAL TAB & 30323 \\
Dementia & Medication & GALANTAMINE 4 MG ORAL TAB & 1232 \\
Dementia & Medication & DONEPEZIL 10 MG ORAL TAB & 31624 \\
Dementia & Medication & DONEPEZIL 5 MG ORAL TAB & 31811 \\
Dementia & Medication & GALANTAMINE 24 MG ORAL C24P & 31774 \\
Dementia & Medication & DONEPEZIL 5 MG ORAL TBDL & 31866 \\
Dementia & Medication & RIVASTIGMINE TARTRATE 3 MG ORAL CAP & 57575 \\
Dementia & Medication & RIVASTIGMINE TARTRATE 6 MG ORAL CAP & 58034 \\
Dementia & Medication & DONEPEZIL ORAL & 60609 \\
Dementia & Medication & GALANTAMINE 8 MG ORAL TAB & 6685 \\
Dementia & Medication & MEMANTINE ORAL & 73925 \\
Dementia & Medication & DONEPEZIL 10 MG ORAL TBDL & 31609 \\
Dementia & Medication & RIVASTIGMINE 4.6 MG/24 HR TRANSDERMAL PT24 & 32859 \\
Dementia & Medication & MEMANTINE 5 MG ORAL TAB & 20149 \\
Dementia & Medication & GALANTAMINE 8 MG ORAL C24P & 31828 \\
Dementia & Medication & MEMANTINE 5–10 MG ORAL DSPK & 20151 \\
Dementia & Medication & RIVASTIGMINE TARTRATE 4.5 MG ORAL CAP & 57679 \\
Dementia & Medication & GALANTAMINE 16 MG ORAL C24P & 31604 \\
Dementia & Medication & RIVASTIGMINE 9.5 MG/24 HR TRANSDERMAL PT24 & 33175 \\
Dementia & Medication & MEMANTINE-DONEPEZIL 28–10 MG ORAL CSPX & 143836 \\
Dementia & Medication & MEMANTINE-DONEPEZIL 7–10 MG ORAL CSPX & 150473 \\
Dementia & Medication & MEMANTINE-DONEPEZIL 14–10 MG ORAL CSPX & 143871 \\
Dementia & Medication & MEMANTINE 28 MG ORAL CSPX & 101254 \\
Dementia & Medication & MEMANTINE-DONEPEZIL 21–10 MG ORAL CSPX & 150139 \\
Dementia & Medication & MEMANTINE 7 MG ORAL CSPX & 112860 \\
Dementia & Medication & MEMANTINE 7–14–21–28 MG ORAL C24K & 101328 \\
Dementia & Medication & DONEPEZIL 23 MG ORAL TAB & 111311 \\
Dementia & Medication & MEMANTINE 14 MG ORAL CSPX & 101112 \\
Dementia & Medication & MEMANTINE 21 MG ORAL CSPX & 101141 \\
Dementia & Medication & RIVASTIGMINE 13.3 MG/24 HR TRANSDERMAL PT24 & 98524 \\
Knee OA & Diagnosis & Bilateral primary osteoarthritis of knee & M17.0 \\
Knee OA & Diagnosis & Unilateral primary osteoarthritis, unspecified knee & M17.10 \\
Knee OA & Diagnosis & Unilateral primary osteoarthritis, right knee & M17.11 \\
Knee OA & Diagnosis & Unilateral primary osteoarthritis, left knee & M17.12 \\
Knee OA & Diagnosis & Bilateral post-traumatic osteoarthritis of knee & M17.2 \\
Knee OA & Diagnosis & Unilateral post-traumatic osteoarthritis, unspecified & M17.30 \\
Knee OA & Diagnosis & Unilateral post-traumatic osteoarthritis, right knee & M17.31 \\
Knee OA & Diagnosis & Unilateral post-traumatic osteoarthritis, left knee & M17.32 \\
Knee OA & Diagnosis & Other bilateral secondary osteoarthritis of knee & M17.4 \\
Knee OA & Diagnosis & Other unilateral secondary osteoarthritis of knee & M17.5 \\
Knee OA & Diagnosis & Osteoarthritis of knee, unspecified & M17.9 \\
Pancreatic Cancer & Diagnosis & Malignant neoplasm of head of pancreas & C25.0 \\
Pancreatic Cancer & Diagnosis & Malignant neoplasm of body of pancreas & C25.1 \\
Pancreatic Cancer & Diagnosis & Malignant neoplasm of tail of pancreas & C25.2 \\
Pancreatic Cancer & Diagnosis & Malignant neoplasm of pancreatic duct & C25.3 \\
Pancreatic Cancer & Diagnosis & Malignant neoplasm of other parts of pancreas & C25.7 \\
Pancreatic Cancer & Diagnosis & Malignant neoplasm of overlapping sites of pancreas & C25.8 \\
Pancreatic Cancer & Diagnosis & Malignant neoplasm of pancreas, unspecified & C25.9 \\
\end{longtable}

\subsection{Hyperparameters}
\label{app:_hyperparam}

\begin{longtable}{ll}
\caption{Training hyperparameters and configuration settings.} \\
\toprule
\textbf{Hyperparameter} & \textbf{Value} \\
\midrule
\endfirsthead

\multicolumn{2}{l}{\textit{(Continued from previous page)}} \\
\toprule
\textbf{Hyperparameter} & \textbf{Value} \\
\midrule
\endhead

\midrule
\multicolumn{2}{r}{\textit{(Continued on next page)}} \\
\endfoot

\bottomrule
\endlastfoot

\texttt{n\_embd} & 2,048 \\
\texttt{n\_head} & 32 \\
\texttt{n\_layer} & 32 \\
\texttt{n\_tokens} & 42,337 \\
\texttt{bias} & false \\
\texttt{dropout} & 0 \\
\texttt{block\_size} & 512 \\
\texttt{batch\_size} & 16 \\
\texttt{optimizer} & AdamW \\
\texttt{beta1} & 0.9 \\
\texttt{beta2} & 0.95 \\
\texttt{decay\_lr} & true \\
\texttt{grad\_clip} & 1 \\
\texttt{gradient\_accumulation\_steps} & 8 \\
\texttt{learning\_rate} & 0.00022 \\
\texttt{lr\_decay\_iters} & 800,000 \\
\texttt{max\_iters} & 810,000 \\
\texttt{min\_lr} & 0.000022 \\
\texttt{rotary} & true \\
\texttt{temporal\_decay} & 0.5 \\
\texttt{warmup\_iters} & 20,000 \\
\texttt{weight\_decay} & 0.01 \\
\end{longtable}

\section{Zero-shot Evaluations of Large Language Models}
\label{app:llm}

The rapid advancement of large language models (LLMs) has introduced new research opportunities in the healthcare domain. To compare and contrast the diagnostic capabilities of our foundation model, we compare its zero-shot performance against selected LLMs on diagnostic prediction tasks.

\subsection{Prompt Engineering}
The EHR dataset used in this study consists exclusively of structured tabular data. To enable diagnostic prediction using LLMs, we first implemented a preprocessing pipeline to convert each patient's sparse medical history into a textual format, with each record laid out in a JSON-like structure. These structured text representations were then paired with a diagnostic prompt that instructed the model to predict ICD codes along with corresponding reasoning within a specified temporal window. The input template is shown below:

\begin{lstlisting}[]
<s>[SYSTEM_PROMPT]
You are a medical forecasting system tasked with predicting future diagnoses
in ICD-10 (International Classification of Diseases, 10th Revision) code format
for patients based on historical medical data.

### Critical Instructions:
  - Make bold and confident predictions about new diagnoses based on historical
    medical data.
  - Do not hesitate to forecast conditions likely to develop due to disease
    progression, comorbidities, and clinical trends.
  - Consider age-related changes, medication effectiveness, lab trends, visit
    patterns, and seasonal health variations.
  - Predict potential diagnoses between {prediction window start date} and {prediction window end date}.
  - Provide output Only in the specified format.
  - Do Not provide any python codes.
  - Always include [END] at the end of the output to signify completion.
</s>

### Input Format:
The input JSON contains the patient's medical history, including:
  - Patient ID
  - Visit timestamps
  - Patient age (if available)
  - ICD-10 codes per visit (if any)
  - Prescribed medications (if any)
  - Laboratory values (if any)

Below is the patient's historical medical data:
{medical records}

### Output Format:
Predict ICD-10 codes between {prediction window start date} and {prediction window end date} in the format below:
<RESPONSE>
```json
{
  "{patient_id}": {
    "Reasoning": "Provide reasoning for the predictions here...",
    "ICD-10 Code": ["ICD_1", "ICD_2", ...]
  }
}
[END]
</RESPONSE>
\end{lstlisting}

Due to the token length limitations of LLMs, records for patients with extensive medical histories often exceeded the maximum input length. To address this, we employed a recursive two-stage prompting strategy. When an input record exceeded the token limit, it was first segmented into fixed-length temporal windows. Each segment was then summarized via a model-generated response using a summarization prompt. The resulting summaries were concatenated and re-evaluated against the token limit. The summarization process was repeated recursively until the record fit within the model’s input constraints. The final diagnosis was generated by applying the original diagnostic template to the refined summary. The summarization prompt used is provided below:

\begin{lstlisting}[]
You are an expert medical assistant. Your task is to summarize the given patient's
electronic health records (EHR) into a concise and structured format. The summary
must preserve the chronological sequence of events and capture key medical details
in a factual and clear manner.

### Instructions:
1. **Summarize** the patient's medical history in a clear, structured, and concise format.
2. **Preserve** the timeline of medical events and ICD-10 codes to reflect the patient's
   health progression.
3. **Start** the summary with: "Summary from [first date] to [last date]".
4. **Limit** the summary to **500 words or less**.
5. **Output** the summary as a **valid JSON object**.
6. **End** the JSON output with the completion marker [END].

### Formatting Rules:
- The output must be in **valid JSON format**.
- Ensure all text is enclosed within **double quotes**.
- The summary should be **provided as a string value within a JSON key**.
- Do **not output Python code** or any additional commentary.
[/SYSTEM_PROMPT]

### Input Format:
The input JSON contains the patient's medical history, including:
- Patient ID
- Visit timestamps
- Patient age (if available)
- ICD-10 codes per visit (if any)
- Prescribed medications (if any)
- Laboratory values (if any)

Below are the patient's records:
{medical records}

### Output Format:
- Provide the output in **structured JSON** format.
- End the output with [END].
- **DO NOT** output python codes.

Example:
<RESPONSE>
```json
{
  "Summary from {records start date} to {records end date}": "Provide the summary..."
}
```
[END]
</RESPONSE>
\end{lstlisting}

\subsection{Evaluation and Results}
The LLM used in our experiment was LLaMA-3.3-70B-Instruct. To ensure a fair comparison with our proposed foundation model, the evaluation was aligned with the task of onset prediction of a specific medical condition and conducted on the same test dataset. For dementia diagnosis over a 5-year prediction horizon, the model achieved a precision of 0.018 and recall of 0.018. These results highlight the limitations of generic instruction-tuned LLMs in accurately diagnosing future disease onset, particularly in specialized medical domains.

\end{document}